\documentclass[conference]{ieeeconf}
  \IEEEoverridecommandlockouts
\usepackage{cite}
\usepackage{amsmath,amssymb,amsfonts}
\usepackage{algorithmic}
\usepackage{graphicx}
\usepackage{placeins}
\usepackage{textcomp}
\usepackage{cite}

\providecommand{\fig}[1]{Figure \ref{fig:#1}}

\providecommand{\figloc}[1]{Figures/#1}

\newcommand{\bm}[1]{\mbox{{\boldmath $#1$}}}

\begin{document}

\title{Resolution-Enhanced MRI-Guided Navigation of Spinal Cellular Injection Robot \\
}

\author{\authorblockN{Daniel Enrique Martinez\authorrefmark{1}, Waiman Meinhold\authorrefmark{1}, John Oshinski\authorrefmark{2}, Ai-Ping Hu\authorrefmark{3}, and Jun Ueda\authorrefmark{4}}
\authorblockA{\authorrefmark{1}Robotics PhD Program,
Georgia Institute of Technology}
\authorblockA{\authorrefmark{2}Department of Radiology,
Emory University}
\authorblockA{\authorrefmark{3}Food Processing Technology Division,
Georgia Tech Research Institute}
\authorblockA{\authorrefmark{4}The George W. Woodruff School of Mechanical Engineering,
Georgia Institute of Technology}
\authorblockA{\authorrefmark{1}Email: dmartinez43@gatech.edu}
\thanks{
This work was partially supported by Georgia Tech Interdisciplinary Institute for Robotics and Intelligent Machines Seed Grant Program. 
This material is also based upon work supported by the National Science Foundation under Grant Nos. 1545287 and 1662029.
Any opinions, findings, and conclusions or recommendations expressed in this material are those of the author(s) and do not necessarily reflect the views of the National Science Foundation.
}
}

\maketitle
\pagenumbering{gobble}


\begin{abstract}
This paper presents a method of navigating a surgical robot beyond the resolution of magnetic resonance imaging (MRI) by using a resolution enhancement technique enabled by high-precision piezoelectric actuation. 
The surgical robot was specifically designed for injecting stem cells into the spinal cord. This particular therapy can be performed in a shorter time by using a MRI-compatible robotic platform than by using a manual needle positioning platform. 
Imaging resolution of fiducial markers attached to the needle guide tubing was enhanced by reconstructing a high-resolution image from multiple images with sub-pixel movements of the robot.  
The parallel-plane direct-drive needle positioning mechanism  positioned the needle guide with a high spatial precision that is two orders of magnitude higher than typical MRI resolution up to 1 mm.
Reconstructed resolution enhanced images were used to navigate the robot precisely that would not have been possible by using standard MRI. 
Experiments were conducted to verify the effectiveness of the proposed enhanced-resolution image-guided intervention. 
\end{abstract}

\section{Introduction}
\label{sec:introduction}

\begin{figure}[t]
 	\centering
 	\includegraphics[width=1.0\columnwidth]{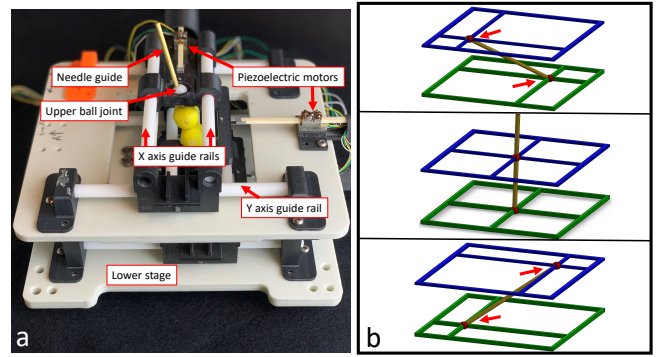} 
 	\caption{The AutoSPINe robot, a) the robot with important components labelled, and b) parallel-plane needle guide positioning concept.}
	\label{fig:Robot} 
\end{figure}

Recent advances in the development of cellular therapeutics for the treatment of both degenerative and acute conditions have created a new need for direct injection into the spinal cord. This procedure can be conducted manually with MR images guiding needle injection, or by a laminectomy. However, these two currently available options result in significant trauma to the region around the spinal cord, or excessive MRI scanner and operating room time, both with questionable accuracy in delivering cells to the targeted locations. 
To address this need, the authors previously reported the development of a highly accurate needle guide positioning robot for use in MRI guided direct spinal cord injection procedures \cite{MeinholdBMES}. Although that system demonstrated accuracy potential up to 14 microns in planar positioning, imaging resolution proved to be a limiting factor in the repeatability of robot motion, as the robot's inherent repeatability was an order of magnitude beyond that of the MRI (1mm). The robot and principle of operation are shown in \fig{Robot}a and \fig{Robot}b respectively. The high repeatability of the robot in comparison to the only other MRI guided spinal injection robot \cite{Squires2018} has the potential to enable novel methods for increasing imaging performance in the MRI.   

Super Resolution (SR) is a method of merging multiple low-resolution images of a scene to estimate a higher resolution image. Several images of the same scene with slightly different imaging conditions, usually spatial shifts, are needed.  An iterative back propagation (IBP) algorithm was first introduced in \cite{irani1993motion} that has been widely used in many applications. 

SR has also been used in robotic applications to improve the capabilities of image sensors. Attempts have been made to apply SR to compensate for limited performance of  low cost and lightweight cameras for the navigation and self-positioning of unmanned flying vehicles \cite{okarma2015application}. Another study used SR to serve as an artificial zoom for minimally invasive surgery, reducing the need for camera motion during surgery due to the enlarged Field of View (FOV) \cite{lerotic2006use}. 

Applications of SR to MRI require a slightly different approach because images are volume slices. In some studies, a spatial shift was introduced by changing the FOV of MRI. FOV shifts have been used to enhance resolution in the inter-slice direction \cite{greenspan2002mri}, but in-plane super resolution is generally regarded to be fundamentally impossible without introducing physical shifts of objects in the scene\cite{plenge2012super}. 
Another study proposed an M-estimation solution 
 to account for the fetal motion and slice thickness that created artifacts in fast MRI techniques needed to capture images of moving fetuses\cite{gholipour2010robust}.  


SR has shown to be a topic of interest applied to both robot control and MRI imaging, but as of yet SR techniques have not been fully applied to visual navigation of robots in MRI. This paper will leverage the spatial information gained from enhanced resolution MRI images to more precisely control an MRI-compatible high-precision needle positioning robot. 

This paper is organized as follows: Section \ref{sec:autopsine} introduces the design and working principles of an MRI-compatible needle positioning robot, the AUTOmated Spinal Precision Injection Needle positioning robot (AutoSPINe).  Section \ref{sec:visualnavigation} describes the concept of visual navigation using SR MRI images. Section \ref{sec:imagingprocedure} explains specific image processing procedures. Sections \ref{sec:numericalanalysis} and \ref{sec:experiments} present results from numerical analysis and experiments, followed by discussion and conclusions.

\section{AutoSPINe}
\label{sec:autopsine}
\subsection{Design}

To achieve both a large range of motion and high degree-of-freedom (DOF) positioning, serial-link mechanisms with a rotary actuator placed at each of the revolute joints are widely used. However, this type of serial chain structure in general tends to lose rigidity and accumulate joint-level positioning errors towards the end-point, resulting in large mechanisms. Backlash and play in gears to drive revolute joints are common issues \cite{patel2019preclinical}. Cable-driven mechanisms  \cite{Hungr2011b} can place actuators on a base structure and make the link mechanism lighter and more compact, but flexibility of cables limits end-point positioning precision. 

In this paper, a direct-drive parallel plane mechanism (D$^2$P$^2$) was adopted as shown in \fig{Robot} \cite{MeinholdBMES}. Each plane is a planar x-y positioning mechanism, positioning a ball joint at the center of the stage, driven by orthogonally located linear actuators as shown in \fig{parallelplane}. The upper and lower ball joints can move independently, controlling 4 DOF of the needle guide. The fifth DOF, i.e., needle depth, is controlled by the surgeon inserting the needle into the cannula. The sixth DOF, i.e., needle rotation, is irrelevant for this procedure. Because the actual distance between the ball joints is dependent on the orientation, the cannula is fixed in the lower joint, while the upper joint allows the cannula to slide through the center of the ball joint. 

The cannula and fiducial markers are of particular importance to accurate positioning of the robot, as any inaccuracies in the fiducial centers will be propagated through to the endpoint position of the needle. The cannula is composed of a 4 mm brass tube with 0.5 mm walls. Fiducials can be attached anywhere on the needle guide or X-Y stages as long as the two marker positions reflect displacements of the linear actuators. 
The fiducials are composed of a spherical cavity filled with Vitamin E, surrounded by a polymer shell. The CAD model of a fiducial is shown in \fig{fiducialfig}a. Upper and lower fiducials are identical. The fiducial shells were printed with polylactic acid (PLA)  on a 3D printer (MakerBot Replicator 2). 
MRI compatibility of this robot was verified and reported in \cite{MeinholdBMES}.

 \begin{figure}[t]
 	\centering
 	\includegraphics[width=1\columnwidth]{\figloc{parallelplane.pdf}} 
 	\caption{X-Y linear stage driven by linear piezoelectric actuators.}
	\label{fig:parallelplane} 
 	\centering
 	\includegraphics[width=0.7\columnwidth]{\figloc{FiducialDiagram.pdf}} 
 	\caption{MRI visible Fiducials, a) CAD model with vitamin E cavity in blue, b) image of fiducials on the cannula.}
	\label{fig:fiducialfig} 
\end{figure}


\subsection{Kinematics}

\begin{figure}[t]
 	\centering
 	\includegraphics[width=0.9\columnwidth]{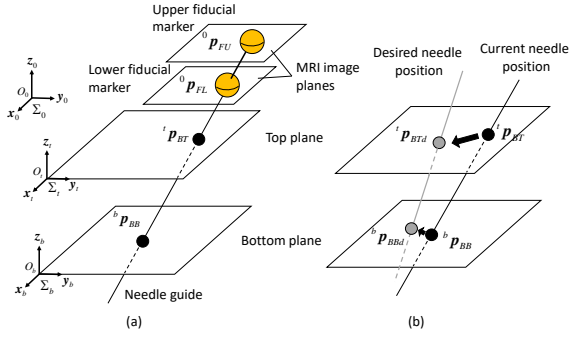} 
 	\caption{AutoSPINe kinematics, a) parallel plane kinematics, b) needle positioning concept}
	\label{fig:kinematics} 
\end{figure}

Forward kinematics is presented to represent the needle guide position in the absolute coordinate frame using ball joint positions in the planar coordinate frames fixed to individual x-y stages. 
Let 
${}^t{\bm p}_{BT}  = \left[ {\begin{array}{c}
   x_t  \\
   y_t  \\
\end{array}} \right]$
be the ball position of the top x-y stage with respect to the coordinate frame fixed to it as shown in \fig{kinematics}. 
Similarity, let ${}^b{\bm p}_{BB}  = \left[ {\begin{array}{c}
   {x_b }  \\
   {y_b }  \\
\end{array}} \right]$ be the ball position of the bottom x-y stage.
Defining ${}^t{\bm P}_{BT}  = \left[ {\begin{array}{*{20}c}
   {{}^t{\bm p}_{BT} }  \\
   1  \\
\end{array}} \right]$
and
${}^b{\bm P}_{BB}  = \left[ {\begin{array}{*{20}c}
   {{}^b{\bm p}_{BB} }  \\
   1  \\
\end{array}} \right]$
, homogeneous transformation,
${}^0{\bm P}_{BT}  = {}^0{\bm T}_t {}^t{\bm P}_{BT} $
and
${}^0{\bm P}_{BB}  = {}^0{\bm T}_t {}^t{\bm P}_{BB} $
provides the ball positions with respect to the base coordinate frame 
$\sum _0 $
where
${}^0{\bm T}_{t}$ and ${}^0{\bm T}_{b}$
are homogeneous transformation matrices. 
Note that without the loss of generality, the x-y planes of $\sum _t $ and $\sum _b $ can be assumed parallel to each other to simplify the kinematic representation.

\subsection{Measurement of needle position using MRI}

Imaging of two fiducial markers  attached to the needle guide enables detection of the 4-DOF position and orientation as shown in \fig{kinematics}a. The spherical exterior of the fiducial markers makes their outer diameter appear as circles in the MRI regardless of their orientation image slice . Running a circle detection algorithm on an image slice with the fiducial marker visible returns the location of the fiducial marker in the image. The 4-DOF position and orientation can then be calculated from the location of the two markers.
Measurement of ${}^0{\bm P}_{FU}$ and ${}^0{\bm P}_{FL}$ determines the line along with the needle guide. The intersection between this line and each of the x-y planes of $\sum _t $ and $\sum _b $ determines the ball joint position. 
 In \fig{Robot} and  \fig{fiducialfig}b, fiducials are placed between the upper and lower stages, while in \fig{kinematics}a they are both placed above the top plane.

\subsection{Image Jacobian and needle positioning}

\begin{figure}[t]
 	\centering
 	\includegraphics[width=0.9\linewidth]{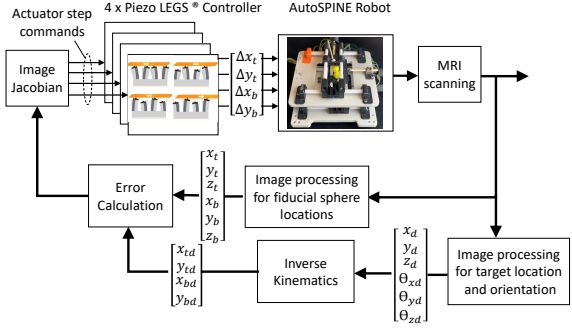} 
 	\caption{MRI guided needle positioning system diagram, linear motor operation diagram adapted from the manufacturer(Micromo, Clearwater FL, USA).}
	\label{fig:needlepositioningdiagram} 
\end{figure}

Image Jacobians relate small actuator displacements, $\Delta x_t, \Delta y_t, \Delta x_b, \Delta y_b$, and resultant ball joint displacements expressed in the global coordinate frame, $\Delta {}^0{\bm P}_{BT}$ and $\Delta {}^0{\bm P}_{BB}$. 
As shown in \fig{kinematics}b, 
matching 
${}^0{\bm P}_{BT}$ and ${}^0{\bm P}_{BB}$
with the ones of the desired needle position, 
${}^0{\bm P}_{BTd}$ and ${}^0{\bm P}_{BBd}$, solves the inverse kinematics. Note that the X-Y stages can be operated independently from each other. The solution is unique as long as ${}^0{\bm P}_{BT}$ and ${}^0{\bm P}_{BB}$ exist, i.e., unless the needle is not completely orthogonal to the parallel planes. In general, image-guided needle positioning is performed in an iterative fashion as illustrated in \fig{needlepositioningdiagram}.

\section{Visual robot navigation with super-resolution MRI}
\label{sec:visualnavigation}

\subsection{Concept}


SR imaging technique processes multiple images with known sub-pixel spatial shifts, typically introduced by moving the camera or the object of interest in the scene, and reconstructs a new image that has a higher resolution than that of the original images. Usually a super-resolution image is produced by numerically solving a cost optimization problem \cite{irani1993motion}. 
When it comes to resolution enhancement of MRI images, there is a technical barrier to introduction of sub-pixel displacements in the target object. This requires highly precise positioning usable in a MRI scanner. 
Note that given the specific architecture of MRI, moving the image acquisition component is not an option. The FOV of the MRI image can be changed, but is generally only considered to be an effective method in improving the resolution in the through-plane direction\cite{plenge2012super}. 




AutoSPINe operates its needle guide and displaces its fiducial markers within MRI imaging resolution of 1mm \cite{MeinholdBMES}. The sub-pixel MRI images are processed to determine the marker positions beyond MRI resolution and navigate the robot toward the target. 
The displacements are introduced by operating one or multiple of PIEZO LEG actuators. Forward kinematics computes resultant small displacements in the fiducial markers. Note that this operation must be performed in an open-loop fashion as sub-pixel movements are essentially not visible in MRI. An arbitrary trajectory may be used to acquire a set of raw images as long as the displacements are known. 

\subsection{Specific imaging procedure}

\label{sec:imagingprocedure}

\subsubsection{Estimation of Image Jacobian}

To compute the image Jacobian three points were used, a central origin point, A 15 mm movement in the x axis, and a 15 mm movement in the y axis. These points were chosen to isolate the x and y actuator movements of the robot if positioned approximately parallel to the image coordinate plane. For the SR and bicubic interpolation (BI) image targeting, we determined that repeating the Jacobian procedure with the larger resolution images yielded the same result as scaling the base resolution Jacobian.

\subsubsection{Super resolution offset matrix generation}

To determine the spatial shifts for image reconstruction, coordinate points are generated randomly ranging from -1 to 1  pixels. The first point is where the first image is taken and is considered the origin point for the following coordinates. The movement in steps required to reach each of the random points is calculated by subtracting each point from the previous point and multiplying the difference by the inverse of the image Jacobian. The generated shifts will be stored in the offset matrix $\bm{M}_k$ for each of $N$ images ($k=1 \cdots N$). 

\subsubsection{Super resolution image construction}

The SR image is reconstructed using gradient descent optimization to minimize the error between the base resolution images, $\bm{I}_k$ ($k=1 \cdots N$), and the current best guess of the high resolution image to iteratively update the best guess of the high resolution image:
$
\hat{\bm{X}} = \mathop {{\rm argmin}}\limits_{\bf X} \left( {\sum\nolimits_{k = 1}^N {\left\| {{\bm D}_k {\bm B}_k {\bm M}_k {\bm X} - {\bm I}_k } \right\|_2^2 } } \right)
$ where 
$\bm{B}_k$ is the blur matrix and $\bm{D}_k$ is the down-sampling matrix. This reconstruction was performed by an example code based on the IBP algorithm \cite{SRCode}. Stopping critera is 100 iterations or when mean square error is below 0.01\% of the mean square error before optimization.

\begin{figure}[t]
 	\centering
 	\includegraphics[width=0.9\columnwidth]{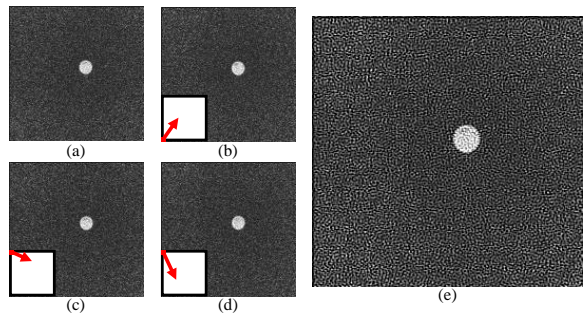} 
 	\caption{MRI SR reconstruction, a) first image with no spatial shift, b-d) three spatially shifted images with arrow representing magnitude and direction of spatial shift relative to first image. box represents 1 pixel e) SR reconstruction.}
	\label{fig:MRISuper} 
\end{figure}

For the experimental validation presented in later sections, four images ($N=4$) were collected to double the resolution of the original images. \fig{MRISuper} shows the four images acquired with a vector arrow representing the spatial shift applied in each image with respect to the first image. After the 4-th image is taken, the AutoSPINe is returned to the first point and the SR image is reconstructed.

\subsubsection{Marker detection}

MATLAB circle finding function, imfindcircles, was used to detect markers in images and MRI image slices. The centerpoint position and radius of the markers were retrieved from this method.

\section{Numerical Analysis}

\label{sec:numericalanalysis}

First, numerical analysis was conducted in MATLAB to quantify the theoretical impact of SR reconstruction of images on the accuracy of the circle finding algorithm previously described. Bicubic Interpolated images were also compared to evaluate the performance of a single image method of increasing resolution. To simulate the loss of information in the image capturing process, an image of a circle was created at a very high resolution (5000$\times$5000 pixels) with a known center point as the ground truth. To create the low resolution image, a blur was applied, the image was down scaled to match the spatial resolution of an MRI image, and noise was added. This was considered the base resolution image.

\begin{figure}[t]
 	\centering
 	\includegraphics[width=1.0\columnwidth]{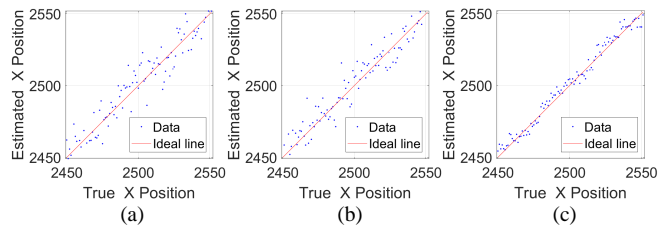} 
 	\caption{Distribution of estimated center point x coordinate plotted against true center point x coordinate, a) shows base resolution distribution, b) shows bicubic interpolation distribution, c) shows SR distribution.}
	\label{fig:numericalanalysis} 
\end{figure}

The process outlined in Table \ref{tab:NumWorkflow} was repeated 100 times and the circle centerpoint coordinates found in the base resolution, interpolated, and SR images were scaled and are plotted against the ground truth centerpoint in Figure \ref{fig:numericalanalysis}.

The mean error of the base resolution images was 0.156 normalized pixels (percentage of image size). Applying bicubic interpolation yielded a similar mean error of 0.135. The SR images yielded a mean error of 0.066, verifying  that reconstructing SR images improves accuracy of object targeting. The improvement from the base resolution images to the SR images, as well as the interpolated images to the SR images was confirmed to be statistically significant with a two-tailed F-test ($p<$ 0.0001).

\begin{table}[t]
\caption{Workflow of Numerical Analysis}
\vspace{-0.5cm}
\label{tab:NumWorkflow}
\begin{center}
\begin{tabular}{|c|l|c|}
\hline
 Step & Description\\
\hline
1 & Create image of very high resolution circle\\
2 & Apply blur, downscale, and add noise to image\\
3 & Find center point of circle in generated Base Resolution Image\\
4 & Use bicubic interpolation to double resolution of Base Image\\
5 & Find centerpoint of circle in Interpolated Image\\
6 & Repeat step 1-  2 to create 3 shifted Base Resolution Images\\
7 & Construct SR image from the 4 BR images and find centerpoint\\
8 & Repeat steps 1 - 7 with the circle shifted 1 pixel to the right\\
\hline
\end{tabular}
\end{center}
\caption{Workflow of Benchtop and MRI Experiments}
\vspace{-0.5cm}
\label{tab:MRIWorkflow}
\begin{center}
\begin{tabular}{|c|l|c|}
\hline
 Step & Description\\
\hline
1 & Take image of robot in central position\\
2 & Move robot and take images to calculate Image Jacobian\\
3 & Create puncture and take image(s) at target position\\
4 & Move robot to random point and take image(s)\\
5 & Calculate error between current and target position\\
6 & Move robot calculated error and take image(s)\\
7 & Repeat steps 5 - 6 until error is less than 1, then create puncture\\
8 & Repeat Steps 4 - 7 for desired number of punctures\\
\hline
\end{tabular}
\end{center}
\end{table}

\section{Experiments}
\label{sec:experiments}

\subsection{Benchtop experiments}
 
To confirm the findings of our numerical analysis applied to robot control, a set of experiments were designed to measure and compare the precision of needle positioning and injection guided by regular images against guidance by SR reconstructed images. To measure the positioning precision with more fidelity than the images, a sharpened rod was pushed through the cannula to puncture a target mounted below the robot. A flat sheet of paper clamped between an acetal resin plate and an ABS plate was used as the target so the punctures would be visible and the distance between punctures could be measured. The AutoSPINe was attached rigidly on top of the target, as seen in Figure \ref{fig:Benchtop}c.

The bottom plane was kept stationary for the experiments to simplify the kinematics of the robot to single plane motion in two axes. This adjustment allows for the replacement of the two fiducials with a red circle printed on a small piece of paper (HP Laserjet 4700dn, 600 dpi) mounted on the top plane. The printed fiducial ensures that there is a clear, flat circle to be found in the images, shown in Figure \ref{fig:Benchtop}b. The benchtop (BT) experiments were designed to be repeated in MRI with minimal adjustments, so an RGB Camera (Intel Realsense) was mounted onto a tripod and placed over the robot, mimicking the coronal view in MRI, as seen in Figure \ref{fig:Benchtop}a.

The camera used for the benchtop experiment has a native resolution of 640$\times$480 pixels. Spatial resolution was set at 1.5 pixels per mm by placing the camera at a height of 33 cm . The images were downscaled using interpolation to match the spatial resolution of the MRI images which is 1 pixel per mm. The images were also cropped to match the FOV of the MRI, 128$\times$128 pixels. An image acquired from this procedure is shown in Figure \ref{fig:Benchtop}b.

The experiments were performed as outlined in Table \ref{tab:MRIWorkflow}. The target position is considered to be the position of the fiducial markers when the first puncture in the trial is made.  For the bicubic interpolation trial, each image was interpolated to double the resolution and used for targeting and updates. The SR trial also follows the same procedure but for each image taken, the SR image construction procedure described in Section \ref{sec:imagingprocedure} was followed to construct the SR image. An example of an SR image acquired from the benchtop experiments is shown in Figure \ref{fig:Benchtop}d. For each trial, the target paper was replaced or moved so the punctures from each trial could be analyzed separately. 
14 punctures were made in each trial to compare the positioning repeatability with the RGB camera base resolution, SR method, and Bicubic Interpolation. 
Each puncture point from the experiments is shown in \fig{Results}a. The distance of each point to the group means were calculated using a stereo microscope (Model
S6D, Leica, Wetzlar, Germany)

A two tailed F-test validates that the variances of the standard and SR groups are statistically different ($ p < .05$). The standard deviations were 0.33 mm and 0.18 mm for the standard and SR targeting groups respectively, with the interpolation-based targeting group producing a 0.23 mm standard deviation. The number of iterations and time needed for each puncture, as well as the puncture precision are summarized in Table \ref{tab:Results}.

 \begin{figure}[t]
 	\centering
 	\includegraphics[width=1.0\columnwidth]{\figloc{BTSetup.jpg}} 
 	\caption{Benchtop experimental set up, b) shows base resolution image acquired from experiment, while c) shows target mounted below AutoSPINe, and d) shows a Super Resolution image reconstructed from
 	benchtop experiments.}
	\label{fig:Benchtop} 
 	\centering
 	\includegraphics[width=1.0\columnwidth]{\figloc{ResultsV2.pdf}} %
 	\caption{Experimental results, base resolution targeting in blue, interpolated targeting in green and super resolution targeting in red, circles represent the standard deviation of each group, a) shows benchtop results, and b) shows MRI results.}
	\label{fig:Results} 
\end{figure}

\subsection{MRI experiments}

\begin{figure}[t]
 	\centering
 	\includegraphics[width=1.0\columnwidth]{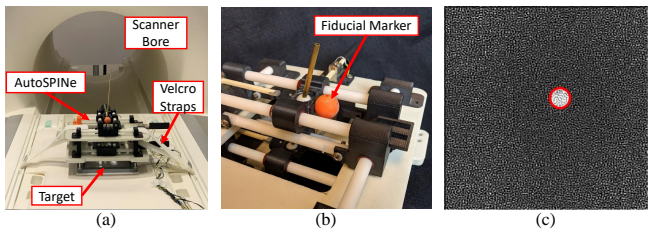} 
 	\caption{MRI experimental set up, b) shows fiducial marker attached to top plane of robot, c) shows a super resolution image reconstructed from MRI experiments with a circle plotted based on image finding algorithm.}
	\label{fig:MRISetup} 
\end{figure}

The MRI experiments were performed similarly to the benchtop experiments, with the AutoSPINe secured rigidly onto the scanner bed as shown in Figure \ref{fig:MRISetup}a. The fiducial marker used for the MRI experiments is shown in Figure \ref{fig:MRISetup}b, mounted on the top stage of the AutoSPINe since only two DOF positioning is considered for the experiments. Only one slice of the MRI images was used to calculate the centerpoint of the fiducial in that plane. A Processed SR image with detected circle overlaid on image is shown in Figure \ref{fig:MRISetup}c.

Three punctures were made in the base resolution experiment and the SR experiment. Experimental positioning results are shown in \fig{Results}b. Scanner time restrictions precluded collection of a statistically significant sample size, however, the standard deviation of the SR group was 33\% smaller than that of the MRI resolution group, 0.60 mm vs. 0.40 mm. The number of iterations and time needed for each puncture are also summarized in Table \ref{tab:Results}.

\section{Discussion}


\subsection{Positioning performance}
This work's primary focus is the utility of SR image reconstruction in the positioning of a robot in the MRI environment. The experimental results detailed in \fig{Results} clearly demonstrate that SR image reconstruction produces more repeatable positioning of the robot. While the MRI results do not rise to the level of statistical significance, they follow the trend shown in benchtop experiments. The results also indicate that the use of bicubic interpolation to ``scale up" a single image for targeting improves performance as well, but not as much as the full SR image reconstruction. 

There is an important difference in the variance of the MRI and benchtop experiments, this is likely due to the lower native resolution of the MRI, and the impact of this lower resolution on the precision of the iterative positioning method. Although the images used in the benchtop experiments were scaled to match the spatial resolution of the MRI images, since interpolation was used to downscale the images, they still provide a more accurate image of the fiducial marker.
This performance improvement is important to the utility of precision MRI positioning robots, as MRI resolution has been a limiting factor in positioning performance.

\begin{table}[t]
\caption{Results of Benchtop and MRI Experiments}
\label{tab:Results}
\vspace{-0.4cm}
\begin{center}
\begin{tabular}{|c|c|c|c|c|c|c|}
\hline
 & BT & \hspace{-0.2cm} BT SR \hspace{-0.2cm} & \hspace{-0.2cm} BT BI \hspace{-0.2cm} & MRI & \hspace{-0.2cm} MRI SR \hspace{-0.2cm} \\
\hline
\hline
Number of iterations   & 1.08 & 1.58 & 1.17 & 2.00 & 1.50\\
 & (0.29) & (0.67) & (0.39) & (1.41) & (0.71)\\
 \hline
Time  (min) & 1.08  & 4.75 & 1.08 & 7.00 & 26.50\\
 & (0.29) & (2.34) & (0.29) & (5.66) & (9.19)\\
\hline
\hspace{-0.2cm} Puncture precision (mm) \hspace{-0.2cm}  & (0.33)  & (0.18)  & (0.23) & (0.60) & (0.40)\\
\hline
\multicolumn{6}{l}{Mean (STD). Precision of needle puncture was evaluated by STD.}
\end{tabular}
\end{center}
\end{table}

\subsection{Imaging time cost for precision}
Although a significant increase in positioning repeatability was found due to the SR based robot navigation method described, this comes with the necessity of increased imaging time. Because the robot used relies on an iterative image feedback method for positioning control, it is likely not necessary to implement the SR reconstruction at each iteration in order to realize a performance improvement. A more efficient scheme would be to do initial positioning in the MRI native resolution, only performing the SR procedure once the robot's positioning is coincident with the target at the precision of the MRI. In other words, SR reconstruction is only necessary once the robot has reached the limit of the MRI resolution.

\section{Conclusion}
This paper evaluated the impact of super resolution reconstruction on visual navigation of a surgical robot in MRI through a numerical analysis, benchtop experiments with an RGB camera, and MRI experiments. The numerical analysis in MATLAB verified the concept that SR images more accurately represent the ground truth of an image as compared to the base resolution images and images interpolated to the same resolution achieved by SR.

Benchtop experiments proved that the improvement in circle finding accuracy translates to improved repeatability when used for robot control. The SR trial had a standard deviation of 0.18 mm, a 45\% improvement over the 0.33 mm deviation of the base resolution trial. MRI experiments also showed an increased precision, with the SR trial having a standard deviation of 0.40 mm, which was 33\% better than the 0.60 mm standard deviation from the base resolution trial. 

It is worth noting that this improvement resulted from doubling the base resolution, and we expect even higher improvements if SR is used to scale the image by a larger factor. These results are promising as they show that SR can be used to exceed the capabilities of imaging hardware for visual navigation at the cost of increased time of image acquisition. Time saving measures will need to be implemented to make SR useful in clinical trials.

\section*{Acknowledgment}
The authors would like to thank Vishwadeep Ahluwalia of the Center for Advanced Brain Imaging (CABI) for his help in the MRI imaging part of this work, as well as Yash Chitalia of the Georgia Tech Robomed Lab for assistance with the measurement of benchtop and MRI results. 

\FloatBarrier{}
\bibliographystyle{unsrt}
\bibliography{superresolution.bib}

\end{document}